\documentclass[10pt,twocolumn,letterpaper]{article}

\usepackage[pagenumbers]{cvpr} % To force page numbers, e.g. for an arXiv version

\usepackage{graphicx}
\usepackage{amsmath}
\usepackage{amssymb}
\usepackage{booktabs}
\usepackage{comment}
\usepackage{caption}
\usepackage{stfloats}
\usepackage{times}
\usepackage{epsfig}
\usepackage{ctable}
\usepackage{colortbl}
\usepackage{lipsum}
\usepackage{xcolor}
\usepackage{stfloats}

\usepackage{multirow}

\usepackage{subcaption}

\usepackage[accsupp]{axessibility}  % Improves PDF

\usepackage[pagebackref,breaklinks,colorlinks]{hyperref}

\usepackage[capitalize]{cleveref}
\crefname{section}{Sec.}{Secs.}
\Crefname{section}{Section}{Sections}
\Crefname{table}{Table}{Tables}
\crefname{table}{Tab.}{Tabs.}

\def\cvprPaperID{2916} % *** Enter the CVPR Paper ID here
\def\confName{CVPR}
\def\confYear{2023}

\newsavebox\CBox
\def\textBF#1{\sbox\CBox{#1}\resizebox{\wd\CBox}{\ht\CBox}{\textbf{#1}}}

\newcommand{\OURS}{Mask3D}
\definecolor{Gray}{gray}{0.92}
\definecolor{darkgreen}{rgb}{0.13, 0.55, 0.13}

\begin{document}

\title{\OURS: Pre-training 2D Vision Transformers by Learning Masked 3D Priors}

\author{%
Ji Hou$^{1}$~~~~~Xiaoliang Dai$^{1}$~~~~~Zijian He$^{1}$~~~~~Angela Dai$^{2}$~~~~~Matthias Nie{\ss}ner$^{2}$ \vspace{0.2cm}\\
$^{1}$Meta Reality Labs~~~~~$^{2}$Technical University of Munich
}

%%%%%%%%% TEASER
\twocolumn[{%
	\renewcommand\twocolumn[1][]{#1}%
	\maketitle
	\begin{center}
		\vspace{-0.35cm}
	\includegraphics[width=1.0\linewidth]{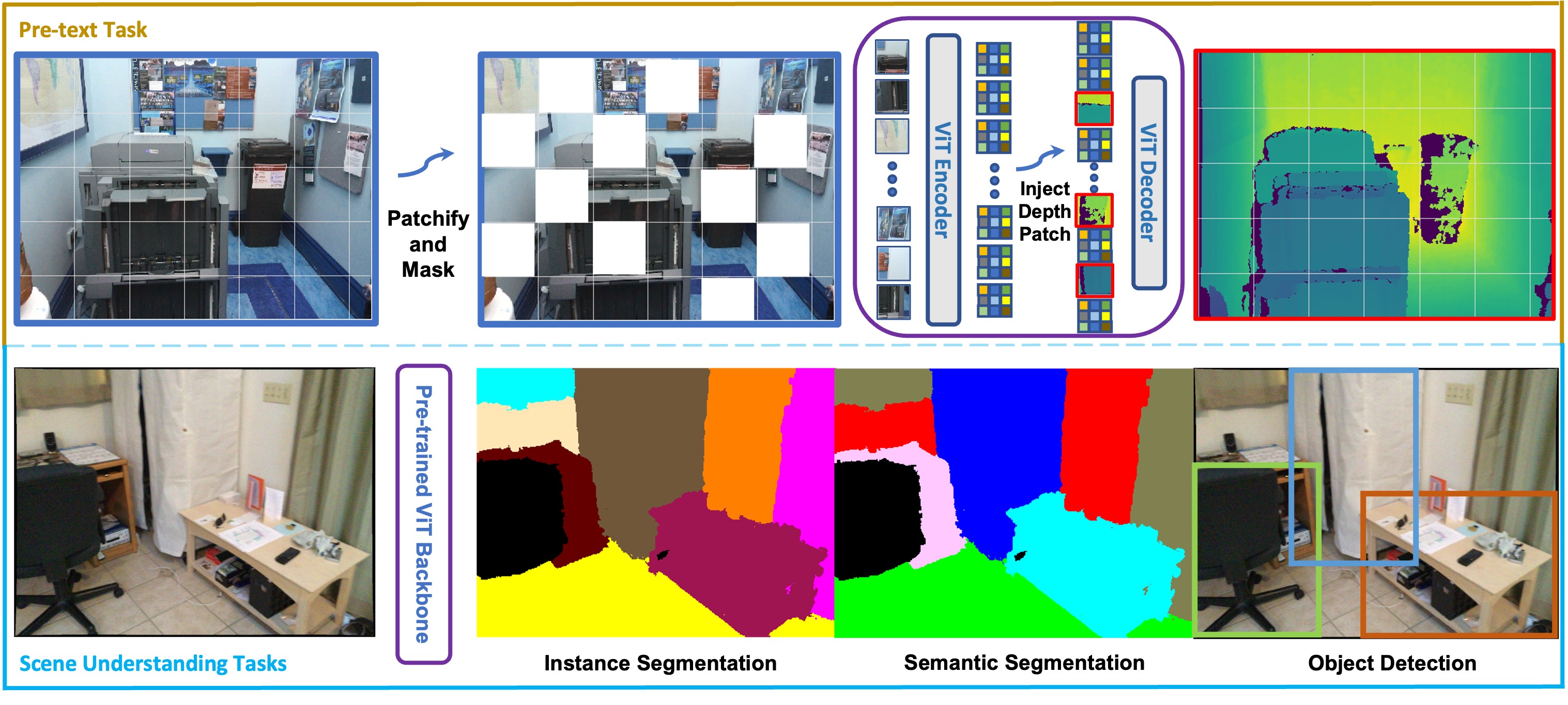}
		\captionof{figure}{
		We present \OURS{}, which learns to embed 3D priors to 2D representations for image understanding tasks, based on a self-supervised pre-training formulation from single RGB-D views, without requiring any camera pose or multi-view correspondence information.
	Our pre-training takes masked RGB and depth patches as input to reconstruct the dense depth map, and the pre-trained color backbone is used to fine-tune various downstream image understanding tasks.
	This results in effective ViT pre-training for a variety of downstream tasks and datasets.
		}
		\vspace{-0.05cm}
		\label{fig:teaser}
	\end{center}
}]

\maketitle

\begin{abstract}
Current popular backbones in computer vision, such as Vision Transformers (ViT) and ResNets are trained to perceive the world from  2D images.
However, to more effectively understand 3D structural priors in 2D backbones, we propose \OURS{} to leverage existing large-scale RGB-D data in a self-supervised pre-training to embed these 3D priors into 2D learned feature representations.
In contrast to traditional 3D contrastive learning paradigms requiring 3D reconstructions or multi-view correspondences, our approach is simple: we formulate a pre-text reconstruction task by masking RGB and depth patches in individual RGB-D frames.
We demonstrate the \OURS{} is particularly effective in embedding 3D priors into the powerful 2D ViT backbone, enabling improved representation learning for various scene understanding tasks, such as semantic segmentation, instance segmentation and object detection. 
Experiments show that \OURS{} notably outperforms existing self-supervised 3D pre-training approaches on ScanNet, NYUv2, and Cityscapes image understanding tasks, with an improvement of +6.5\% mIoU against the state-of-the-art Pri3D on ScanNet image semantic segmentation.

\end{abstract}
\section{Introduction}
\label{sec:intro}
Recent years have seen remarkable advances in 2D image understanding as well as 3D scene understanding, although their representation learning has generally been treated separately.
Powerful 2D architectures such as ResNets~\cite{he2016deep} and Vision Transformers (ViT)~\cite{dosovitskiy2020image} have achieved notable success in various 2D recognition and segmentation tasks, but focus on learning from 2D image data. Current large-scale RGB-D datasets \cite{dai2017scannet,armeni_cvpr16,chang2017matterport3d,song2015sun,silberman2012indoor} provide an opportunity to learn key geometric and structural priors to provide more informed reasoning about the scale and circumvent view-dependent effects, which can provide more efficient representation learning. In 3D, various successful methods have been leveraging the RGB-D datasets for constrastive point discrimination \cite{xie2020pointcontrast,hou2020exploring,zhang2021self,chen20214dcontrast} for downstream 3D tasks, including high-level scene understanding tasks as well as low-level point matching tasks~\cite{el2021bootstrap,zhang2022pcr}. However, the other direction from 3D to 2D is less explored.

We thus aim to embed such 3D priors into 2D backbones to effectively learn the structural and geometric priors underlying the 3D scenes captured in 2D image projections. Recently, Pri3D~\cite{hou2021pri3d} adopted similar multi-view and reconstruction-based constraints to induce 3D priors in learned 2D representations. However, this relies on not only acquiring RGB-D frame data but also the robust registration of multiple views to obtain camera pose information for each frame. Instead, we consider how to effectively learn such geometric priors from only single-view RGB-D data in a more broadly applicable setting for 3D-based pre-training.

We thus propose \OURS{}, which learns effective 3D priors for 2D backbones in a self-supervised fashion by pre-training with single-view RGB-D frame data. We propose a pre-text reconstruction task to reconstruct the depth map by masking different random RGB and depth patches of an input frame. These masked input RGB and depth are encoded simultaneously in separate encoding branches and decoded to reconstruct the dense depth map.
This imbues 3D priors into the RGB backbone which can then be used for fine-tuning downstream image based scene understanding tasks.

In particular, our self-supervised approach to embedding 3D priors from single-view RGB-D data to 2D learned features is not only more generally applicable, but we also demonstrate that it is particularly effective for pre-training vision transformers.
Our experiments demonstrate the effectiveness of Mask3D on a variety of datasets and image understanding tasks.
We pre-train on ScanNet~\cite{dai2017scannet} with our masked 3D pre-training paradigm and fine-tune for 2D semantic segmentation, instance segmentation, and object detection.
This enables notable improvements not only on ScanNet data but also generalizes to NYUv2~\cite{silberman2012indoor} and even Cityscapes~\cite{cordts2016cityscapes} data.
We believe that \OURS{} makes an important step to shed light on the paradigm of incorporating 3D representation learning to powerful 2D backbones.

In summary, our contributions are:
\begin{itemize}
\vspace{-0.1cm}
    \item We introduce a self-supervised pre-training approach to learn masked 3D priors for 2D image understanding tasks based on learning from only single-view RGB-D data, without requiring any camera pose or 3D reconstruction information, and thus enabling more general applicability.
	\item We demonstrate that our masked depth reconstruction pre-training is particularly effective for the modern, powerful ViT architecture, across a variety of datasets and image understanding tasks.
\end{itemize}

\section{Related Work}

\paragraph{Pre-training in Visual Transformers.} Recently, visual transformers have revolutionized computer vision and attracted wide attention. 
In contrast to popular CNNs that operate in a sliding window fashion, Vision Transformers (ViT) describe the image as patches of 16x16 pixels. 
The Swin Trasnformer~\cite{liu2021swin} has set new records with its hierarchical transformer formulation on major vision benchmarks.
The dominance of visual transformers in many vision tasks has inspired study into how to pre-training such backbones. MoCoV3~\cite{chen2021empirical} first investigated the effects of several fundamental components for self-supervised ViT training. MAE~\cite{he2021masked} then proposed an approach inspired by BERT~\cite{devlin2018bert}, which randomly masks words in sentences and leveraged masked image reconstruction for self-supervised pre-training that achieved state-of-the-art results in ViT. 
A similar self-supervision has also been proposed by MaskFeat~\cite{wei2021masked} for self-supervised video pre-training. MaskFeat randomly masks out pixels of the input sequence and then predicts the Oriented Gradients (HOG) of the masked regions. However, such ViT pre-training methods focus on image or video data, without exploring how 3D priors can potentially be exploited. MultiMAE~\cite{bachmann2022multimae} on the other hand introduces depth priors. However, it requires depth as input not only in pre-training but also in downstream tasks. In addition to depth, human annotations (e.g., semantics) are also leveraged in the pre-training. To achieve a self-supervised pre-training, we do not use semantics in the pre-training and only use RGB images as input in downstream tasks.

\vspace{-0.35cm}
\paragraph{RGB-D Scene Understanding.}
Research in 3D scene understanding have been spurred forward with the introduction of larger-scale, annotated real-world RGB-D datasets \cite{armeni_cvpr16,dai2017scannet,chang2017matterport3d}. This has enabled data-driven semantic understanding of 3D reconstructed environments, where we have now seen notable progress, such as for 3D semantic segmentation~\cite{qi2017pointnet,qi2017pointnetplusplus,dai20183dmv,thomas2019kpconv,choy20194d,graham20183d}, object detection~\cite{voteNet,zhang2020h3dnet,nie2020rfd}, instance segmentation~\cite{hou20193d,yi2018gspn,yang2019learning,lahoud20193d,hou2020revealnet,engelmann20203d,han2020occuseg,jiang2020pointgroup}, and recently panoptic segmentation~\cite{dahnert2021panoptic}. Such 3D scene understanding tasks have been analogously defined to 2D image understanding, which considers RGB-only input without depth information. However, learning from 3D enables geometric reasoning without requiring learning view-dependent effects or resolving depth/scale ambiguity that must be learned when considering solely 2D data. We thus take advantage of existing large-scale RGB-D data to explore how to effectively embed 3D priors for better representation learning for 2D scene understanding tasks.

\vspace{-0.35cm}
\paragraph{Embedding 3D Priors in 2D Backbones.} 
Learning cross-modality features has been seen in extensive studies of the ties between languages and images. In particular, CLIP~\cite{radford2021learning} learns visual features from language supervision during pre-training, showing promising results in zero-shot learning for image classification. Pri3D~\cite{hou2021pri3d} explores 3D-based pre-training for image-based tasks by leveraging multi-view consistency and 2D-3D correspondence with contrastive learning to embed 3D priors into ResNet backbones. 
This results in enhanced features over ImageNet pre-training on 2D scene understanding tasks. However, Pri3D requires camera pose registration across RGB-D video sequences and is specifically designed for CNNs-based architectures. In contrast, we formulate a self-supervised pre-training that operates on only single-view RGB-D frames and leverages masked 3D priors that can effectively pre-train powerful ViT backbones.

\section{Method}

We introduce \OURS{} to embed 3D priors into learned 2D representations by self-supervised pre-training from only single-view RGB-D frames. To effectively learn 3D structural priors without requiring any camera pose information or multi-view constraints, we formulate a pre-text depth reconstruction task to inform the RGB feature extraction to be geometrically aware.
Randomly masked color and depth images are used as input to reconstruct the dense depth map, and the RGB backbone can then be used to fine-tune downstream image understanding tasks. In particular, we show in Sec.~\ref{sec:results} that this single-frame self-supervision is particularly well-suited for powerful vision transformer (ViT) backbones, even without any multi-view information.

\subsection{Learning Masked 3D Priors}
\label{sec:our_method}

\begin{figure*}[h!]
	\centering
	\includegraphics[width=1.0\linewidth]{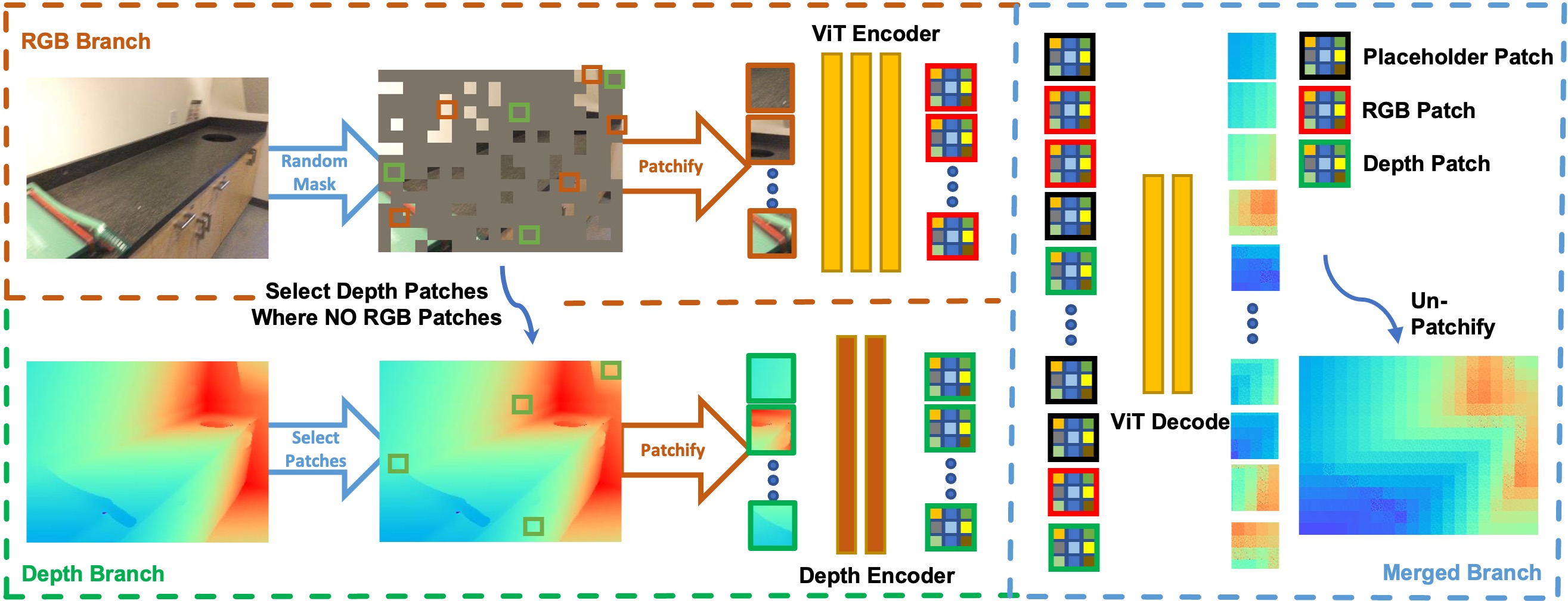}
	\vspace{-0.5cm}
	\caption{\textbf{Overview of Mask3D Pre-training.}
	As a pretext task, we predict dense depth from color and sparse depth signals. 
	We use masked input by randomly selecting a set of patches from the input color image, which are then mapped to higher dimensional feature vectors; input depth is similarly masked and encoded. The color and depth features are then fused into a bottleneck from which dense depth is reconstructed as a self-supervised loss.
	}
	\label{fig:method_overview}
	\vspace{-0.35cm}
\end{figure*}

We propose to learn masked 3D priors to embed to learned 2D backbones by pre-training to reconstruct dense depth from RGB images with the guidance of sparse depth.
That is, for an RGB-D frame $F=(C,D)$ with RGB image $C$ and depth map $D$, we train to reconstruct $D$ from masked patches of $C$ guided with sparse masked patches of $D$. An overview of our approach is shown in Fig.~\ref{fig:method_overview}.

To create masked color and depth $M_c$ and $M_d$ from $C$ and $D$ as input for reconstruction, a 240x320 RGB image $C$ is uniformly divided into 300 16x16 patches, from which we randomly keep a percentage $p_c$ of patches, masking out the others, to obtain $M_c$.
$M_d$ is created similarly by keeping only a percentage $p_d$ of patches, such that the resulting depth patches do not coincide with the RGB patches in $M_c$.

We then train color and depth encoders $\Psi_c$ and $\Psi_d$ to separately encode RGB and depth signals.
RGB patches are fed into $\Psi_c$ and concatenated with a positional embedding, following the ViT architecture, and similarly for depth. 
The positional embedding used encodes the patch location by a cosine function.
Patches and their positional embeddings are then mapped into higher dimensional feature vectors via $\Psi_c$ and $\Psi_d$. The encoders $\Psi_c$ and $\Psi_d$ are built by blocks composed of linear and norm layers. The features from $\Psi_c$ and $\Psi_d$ are then fused in the bottleneck; since depth patches were selected in regions where no RGB patches were selected, there are no duplicate patches representing the same patch location.

For those regions which do not have any associated RGB or depth patch, we use patches of constant values as mask tokens to create a placeholder in the bottleneck to enable reconstructing dense depth at the original image resolution. In the bottleneck, the RGB and depth patch feature vectors, along with the mask tokens, form the input to the decoder. This formulates a reconstruction task from sparse RGB and depth; the joint RGB-D pre-training enables reconstruction from very sparse input, as shown by our ablation on the masked input rations in Sec.~\ref{sec:ablation}. Note that the depth encoder is trained only during pre-training, and only the color ViT encoder (and decoder, if applicable) are used for downstream fine-tuning.

\begin{figure}
\centering
	\centering
	\includegraphics[width=1.0\linewidth]{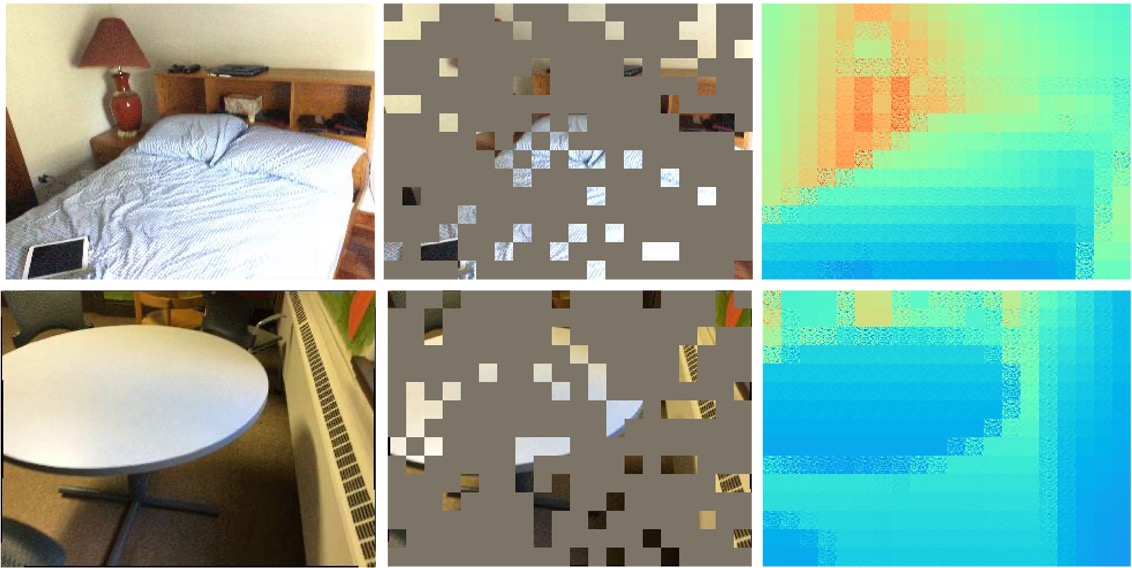}
	\vspace{-0.55cm}
	\caption{\textbf{Pre-trained ViT learns 3D structual priors.} Our proposed pre-training method learns spatial structures from heavily masked RGB images.}
	\label{fig:dense_depth_recon}
	\vspace{-0.2cm}
\end{figure}

To demonstrate the effectiveness of the pre-training task, we demonstrate the depth completion results from the pre-training phase in Fig.~\ref{fig:dense_depth_recon}. A detailed analysis of masking different ratios of color and depth signals is shown in Sec.~\ref{sec:ablation}.

\vspace{-0.5cm}
\paragraph{Pre-training Loss} 
In contrast to the widely used contrastive loss in 3D representation learning, we train for dense depth reconstruction with an $\ell_2$ reconstruction loss.
Similar to MAE~\cite{he2021masked}, we normalize the output patches as well as the target patches prior to computing the loss, which we found to empirically improve the performance.

\section{Results}
\label{sec:results}

We demonstrate the effectiveness of \OURS{} pre-training for ViT~\cite{dosovitskiy2020image} backbones  on semantic segmentation, instance segmentation, and object detection tasks. We pre-train on ScanNet~\cite{dai2017scannet} data and demonstrate the effectiveness of learned masked 3D priors for not only ScanNet downstream tasks but also their transferability to NYUv2~\cite{silberman2012indoor} and even across the indoor/outdoor domain gap to Cityscapes~\cite{cordts2016cityscapes} data.

\subsection{Experimental Setup}
\label{sec:setup}
In this section, we introduce the pre-training and fine-tuning procedures. Our method uses a two-stage pre-training design introduced in the following.

\vspace{-0.25cm}
\paragraph{Stage-I: \OURS{} Encoder Initialization.}
We initialize the RGB encoder with network weights trained on ImageNet~\cite{deng2009imagenet} (as pre-training for our pre-training). To maintain a fully self-supervised pre-training paradigm, we initialize with weights obtained by self-supervised ImageNet pre-training~\cite{he2021masked}. 

\vspace{-0.25cm}
\paragraph{Stage-II: \OURS{} Pre-training on ScanNet.}
\OURS{} pre-training leverages 3D priors in RGB-D frame data, for which we use the color and depth maps of ScanNet~\cite{dai2017scannet}. Note that this does not use any semantic or reconstruction information during pre-training. ScanNet contains 2.5M RGB-D frames from 1513 ScanNet train video sequences. We regularly sample every $25^{\textrm{th}}$ frame without any other filtering (e.g., no control on viewpoint variation).

\vspace{-0.25cm}
\paragraph{Downstream Fine-tuning.} We evaluate our \OURS{} pre-training by fine-tuning a variety of downstream image understanding tasks (semantic segmentation, instance segmentation, object detection).
We consider in-domain transfers on ScanNet image understanding, and further evaluate the out-of-domain transfer on datasets with different statistical characteristics: the indoor image data of NYUv2~\cite{silberman2012indoor}, as well as across a strong domain gap to the outdoor image data of Cityscapes~\cite{cordts2016cityscapes}. For semantic segmentation tasks, we use both encoder and decoder pre-trained with \OURS{}, and for instance segmentation and detection, only the backbone encoder is pre-trained.

\vspace{-0.35cm}
\paragraph{Baselines.} 
To evaluate the effectiveness of our learned masked 3D priors for 2D representations, we benchmark our method against relevant baselines:

\emph{Supervised ImageNet Pre-training (supIN).} This uses the pre-trained weights from ImageNet, provided by torchvision, as is commonly used for image understanding tasks. Here, only ImageNet data is used, and no ScanNet data is involved in the pre-training phase.

\emph{2-Stage MoCoV2 (MoCoV2-supIN$\rightarrow$SN).} Supervised ImageNet pre-trained (supIN) weights are used as network initialization for pre-training. MoCoV2~\cite{he2020momentum} is used for pre-training with randomly shuffled ScanNet images. Here, both ImageNet and ScanNet image data are used without any geometric priors.
 
\emph{2-Stage MAE (MAE-unsupIN$\rightarrow$SN).} Self-supervised ImageNet pre-trained weights are used as network initialization for pre-training. 
MAE~\cite{he2021masked} is used for pre-training with randomly shuffled ScanNet images. Here, both ImageNet and ScanNet image data are used without any geometric priors.

\emph{Pri3D}~\cite{hou2021pri3d}. Supervised ImageNet pre-trained are used to initialize Pri3D pre-training, which leverages multi-view and reconstruction constraints from ScanNet data under a contrastive loss.
Here, both ImageNet and ScanNet data are used, incorporating 3D priors from reconstructed RGB-D video sequences for pre-training.

\begin{table*}[h]
\small
  \centering
  \begin{tabular}{|l|c|c|c|}
  \hline
Pre-training Method ~~~~~~~~~~&~~~~~~~Backbone~~~~~~~&~~~~~~~Pre-training Data~~~~~~~&  ~~~~~~~~~~mIoU~~~~~~~~~~~~ \\
  \hline
  \textcolor{gray}{Scratch}~&\textcolor{gray}{ResNet-50}&\textcolor{gray}{None}&~\textcolor{gray}{39.1}~~~~\\
  ImageNet Pre-training (supIN)~&ResNet-50 & ImageNet             &~~55.7~~~~~\\
  Supervised Pre-training~& ViT & ImageNet+ScanNet             &~~~~~~~~65.9~~\tiny{\textBF{\textcolor{gray}{($+$10.2)}}}~~~\\ \hline
MoCoV2-supIN$\rightarrow$SN~\cite{he2020momentum} ~&ResNet-50 & ImageNet+ScanNet     &~~~~~~~56.6~~\tiny{\textBF{\textcolor{gray}{($+$0.9)}}}~~ \\ 
   Pri3D~\cite{hou2021pri3d}~&ResNet-50&ImageNet+ScanNet&~~~~~~60.2~~\tiny{\textBF{\textcolor{gray}{($+$4.5)}}}~~\\ 
   Pri3D~\cite{hou2021pri3d}~&ViT&ImageNet+ScanNet&~~~~~~59.3~~\tiny{\textBF{\textcolor{gray}{($+$3.6)}}}~~\\
   DINO~\cite{caron2021emerging}~&ViT&ImageNet+ScanNet&~~~~~~58.1~~\tiny{\textBF{\textcolor{gray}{($+$3.6)}}}~~\\ 
   MAE-unsupIN$\rightarrow$SN~\cite{he2021masked}~&ViT&ImageNet+ScanNet&~~~~~~63.3~~\tiny{\textBF{\textcolor{gray}{($+$7.6)}}}~~\\ \hline 
   Ours -- \OURS{} (DINO) ~&ViT&ImageNet+ScanNet&~~~~~~60.5~~\tiny{\textBF{\textcolor{gray}{($+$4.8)}}}~~\\
   Ours -- \OURS{} (MAE) ~&ViT&ImageNet+ScanNet&~~~~~~~\textBF{66.7}~~\tiny{\textBF{\textcolor{darkgreen}{($+$11.0)}}}~~\\ \hline
  \end{tabular}
  \vspace{-0.1cm}
  \caption{\textbf{ScanNet 2D Semantic Segmentation.} \OURS{} significantly outperforms Pri3D as well as other state-of-the-art pre-training approaches that leverage both ImageNet and ScanNet data.
  }
  \vspace{-0.25cm}
\label{tab:scannet_semseg}
\end{table*}

\vspace{-0.35cm}
\paragraph{Implementation Details.} 
\label{sec:implementation}
We use a ViT-B backbone to train our approach.
For pre-training, we use an SGD optimizer with a learning rate of 0.1 and an effective batch size of 128 (accumulated gradients from an actual batch size of 64). The learning rate is decreased by a factor of 0.99 every 1000 steps, and our method is trained for 100 epochs. Fine-tuning on semantic segmentation is trained with a batch size of 8 for 80 epochs. The initial learning rate is 0.01, with polynomial decay with a power of 0.9. Fine-tuning on detection and instance segmentation is trained using Detectron2~\cite{wu2019detectron2} with the 1x schedule. Pre-training experiments are conducted on a single NVIDIA A6000 GPU, or 2 NVIDIA RTX3090 GPUs, or 4 NVIDIA RTX2080Ti GPUs; semantic segmentation experiments are conducted on a single NVIDIA A6000 GPU; instance segmentation and detection experiments are conducted on 8 V100 GPUs.

\begin{table}
\small
  \centering
  \begin{tabular}{|l|ccc|}
  \hline
   Pre-training Method &  AP@0.5 & AP@0.75  & AP~~~~~  \\ \hline
   \textcolor{gray}{Scratch}&\textcolor{gray}{32.7}~~~~~~~&\textcolor{gray}{17.7}~~~~~~~&\textcolor{gray}{16.9}~~~~~~~\\
   ImageNet Pretrain (supIN)&41.7~~~~~~~&25.9~~~~~~~&25.1~~~~~~~\\ \hline
   MoCoV2-supIN$\rightarrow$SN\cite{he2020momentum}&43.5~\tiny{\textBF{\textcolor{gray}{(+1.8)}}}&26.8~\tiny{\textBF{\textcolor{gray}{(+0.9)}}}&25.8~\tiny{\textBF{\textcolor{gray}{(+0.7)}}}\\
   Pri3D~\cite{hou2021pri3d}&43.7~\tiny{\textBF{\textcolor{gray}{(+2.0)}}}&27.0~\tiny{\textBF{\textcolor{gray}{(+1.1)}}}&26.3~\tiny{\textBF{\textcolor{gray}{(+1.2)}}}\\
   MAE-unsupIN$\rightarrow$SN~\cite{he2021masked}&46.1~\tiny{\textBF{\textcolor{gray}{(+4.4)}}}&32.7~\tiny{\textBF{\textcolor{gray}{(+6.8)}}}&30.5~\tiny{\textBF{\textcolor{gray}{(+5.4)}}}\\ \hline 
   \OURS{} (Ours)&\textBF{50.4}~\tiny{\textBF{\textcolor{darkgreen}{(+8.7)}}}&\textBF{35.3}~\tiny{\textBF{\textcolor{darkgreen}{(+9.4)}}}&\textBF{32.7}~\tiny{\textBF{\textcolor{darkgreen}{(+7.6)}}}\\ \hline
  \end{tabular}
\vspace{-0.15cm}
  \caption{\textbf{ScanNet 2D Object Detection.} Fine-tuning with \OURS{} pre-trained models leads to improved object detection results across different metrics, in comparison to ImageNet pre-training, MoCo-style pre-training, and a strong MAE-style pre-training method.
  }  
\vspace{-0.2cm}
\label{tab:scannet_det}
\end{table}

\subsection{ScanNet Downstream Tasks}
\label{sec:scannet}
We demonstrate the effectiveness of representation learning with 3D priors via downstream tasks on ScanNet~\cite{dai2017scannet} images. For fine-tuning, we follow the standard protocol of the ScanNet benchmark~\cite{dai2017scannet} and sample every $100^{\textrm{th}}$ frame, resulting in 20,000 train images and 5,000 validation images. 

\begin{figure*}
	\centering
	\includegraphics[width=0.95\linewidth]{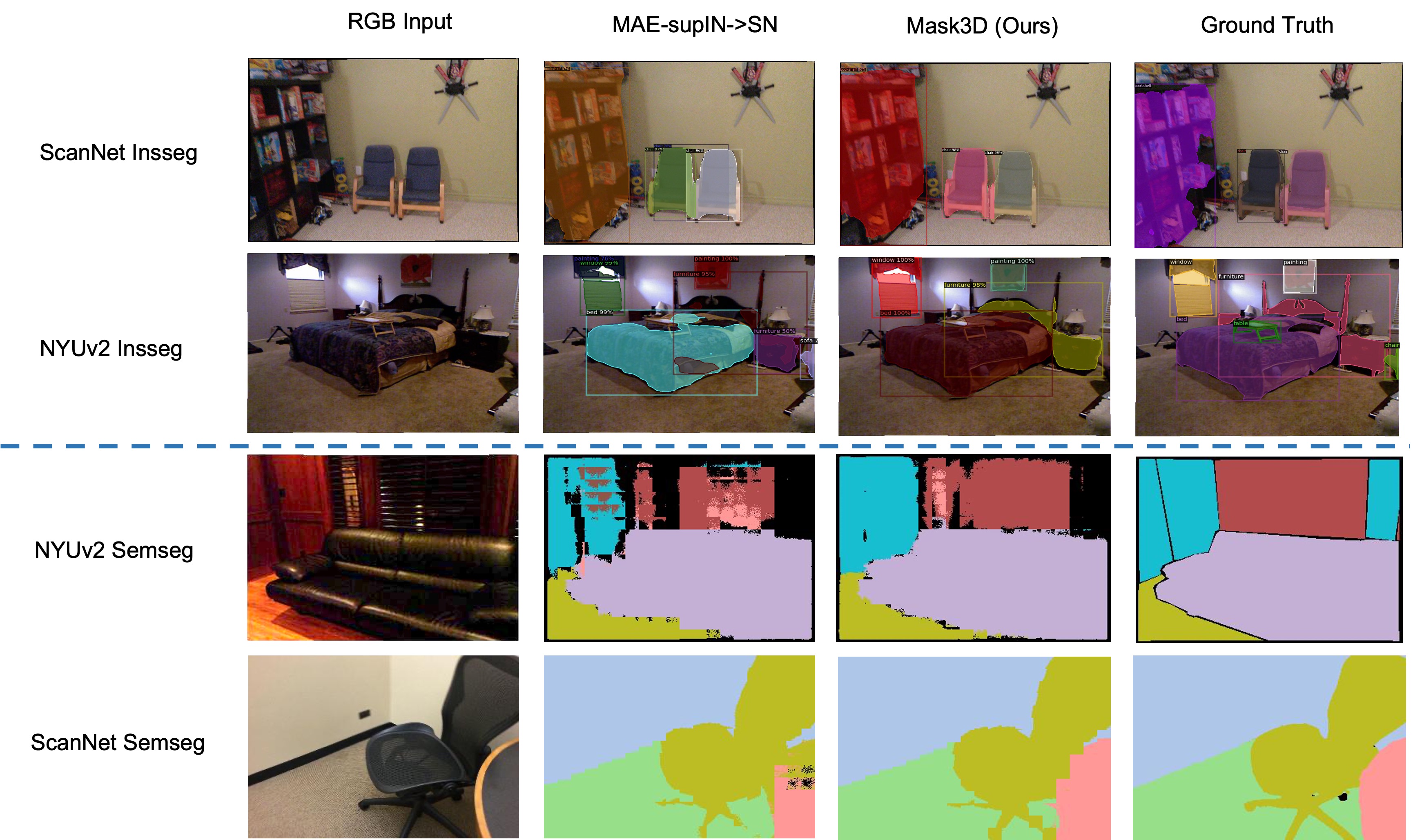}
\vspace{-0.15cm}
	\caption{\textbf{Qualitative Results on Various Tasks across Different Benchmarks.} We visualize predictions on different tasks across various scene understanding benchmarks. From top to bottom rows: instance segmentation on ScanNet, instance segmentation on NYUv2, semantic segmentation on NYUv2, and semantic segmentation results in ScanNet.
	}
\vspace{-0.15cm}
	\label{fig:visualization}
\end{figure*}

\begin{figure*}
	\centering
	\includegraphics[width=1.0\linewidth]{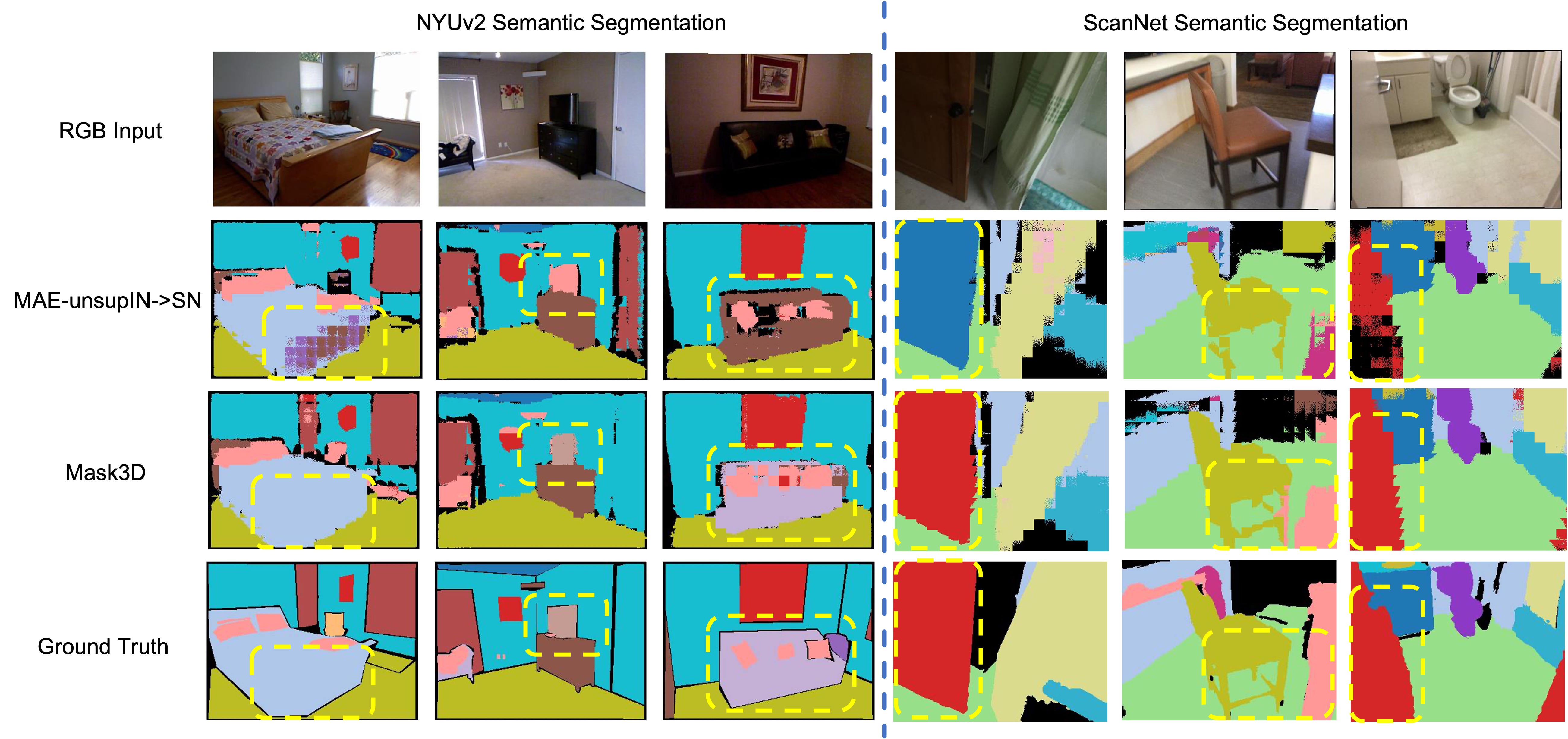}
\vspace{-0.65cm}
	\caption{\textbf{More Qualitative Results on Semantic Segmentation.} We visualize semantic segmentation predictions on various scene understanding benchmarks including ScanNet and NYUv2.
	}
	\label{fig:visualization}
\vspace{-0.35cm}
\end{figure*}

\begin{table}
\small
  \centering
  \begin{tabular}{|l|ccc|}
  \hline
   Pre-training Method &  AP@0.5 & AP@0.75& AP~~~~~ \\ \hline
   \textcolor{gray}{Scratch}&\textcolor{gray}{25.8}~~~~~~~&\textcolor{gray}{13.1}~~~~~~~&\textcolor{gray}{12.2}~~~~~~~\\
   ImageNet Pretrain (supIN)&32.6~~~~~~~&17.8~~~~~~~&17.6~~~~~~~\\ \hline
   MoCoV2-supIN$\rightarrow$SN~\cite{he2020momentum}&33.9~\tiny{\textBF{\textcolor{gray}{(+1.3)}}}&18.1~\tiny{\textBF{\textcolor{gray}{(+0.3)}}}&18.3~\tiny{\textBF{\textcolor{gray}{(+0.7)}}}\\
  Pri3D~\cite{hou2021pri3d}&34.3~\tiny{\textBF{\textcolor{gray}{(+1.7)}}}&18.7~\tiny{\textBF{\textcolor{gray}{(+0.9)}}}&18.3~\tiny{\textBF{\textcolor{gray}{(+0.7)}}}\\
    MAE-unsupIN$\rightarrow$SN~\cite{he2021masked}&37.4~\tiny{\textBF{\textcolor{gray}{(+4.8)}}}&20.3~\tiny{\textBF{\textcolor{gray}{(+2.5)}}}&20.7~\tiny{\textBF{\textcolor{gray}{(+3.1)}}}\\ \hline 
   \OURS{} (Ours)&\textBF{41.2}~\tiny{\textBF{\textcolor{darkgreen}{(+8.6)}}}&\textBF{22.7}~\tiny{\textBF{\textcolor{darkgreen}{(+4.9)}}}&\textBF{22.8}~\tiny{\textBF{\textcolor{darkgreen}{(+5.2)}}}\\ \hline
  \end{tabular}
\vspace{-0.15cm}
  \caption{\textbf{ScanNet 2D Instance Segmentation.}  Fine-tuning with \OURS{} pre-trained models leads to improved instance segmentation results across different metrics compared to ImageNet pre-training, MoCo-style pre-training, and a strong MAE-style pre-training method. 
  }  
\label{tab:scannet_insseg}
\vspace{-.2cm}
\end{table}

\begin{table}
\small
  \centering
  \begin{tabular}{|l|ccc|}
  \hline
   Pre-training Method & AP@0.5 & AP@0.75  & AP~~~~~~~  \\ \hline
   \textcolor{gray}{Scratch}&\textcolor{gray}{17.2}~~~~~~~&\textcolor{gray}{9.2}~~~~~~~&\textcolor{gray}{8.8}~~~~~~~\\
   ImageNet Pretrain (supIN)&25.1~~~~~~~&13.9~~~~~~~&13.4~~~~~~~\\ \hline
   MoCoV2-supIN$\rightarrow$SN~\cite{he2020momentum}&27.2~\tiny{\textBF{\textcolor{gray}{(+2.1)}}}&14.7~\tiny{\textBF{\textcolor{gray}{(+0.2)}}}&14.8~\tiny{\textBF{\textcolor{gray}{(+1.4)}}}\\
   Pri3D~\cite{hou2021pri3d}&28.1~\tiny{\textBF{\textcolor{gray}{(+3.0)}}}&15.7~\tiny{\textBF{\textcolor{gray}{(+1.8)}}}&15.7~\tiny{\textBF{\textcolor{gray}{(+2.3)}}}\\
   MAE-unsupIN$\rightarrow$SN~\cite{he2021masked}&33.6~\tiny{\textBF{\textcolor{gray}{(+8.5)}}}&19.0~\tiny{\textBF{\textcolor{gray}{(+5.1)}}}&19.0~\tiny{\textBF{\textcolor{gray}{(+5.6)}}}\\ \hline 
   \OURS{} (Ours)&\textBF{37.0}~\tiny{\textBF{\textcolor{darkgreen}{(+11.9)}}}&\textBF{21.6}~\tiny{\textBF{\textcolor{darkgreen}{(+7.7)}}}&\textBF{21.3}~\tiny{\textBF{\textcolor{darkgreen}{(+7.9)}}}\\ \hline
  \end{tabular}
  \vspace{-0.15cm}
  \caption{\textbf{NYUv2 2D Instance Segmentation.}  Fine-tuning with \OURS{} pre-trained models leads to improved instance segmentation results across different metrics compared to previous methods, demonstrating the cross-dataset transfer ability of \OURS{}.
  }
  \vspace{-0.35cm}
\label{tab:nyu_insseg}
\end{table}

\vspace{-0.35cm}
\paragraph{2D Semantic Segmentation.} 
Tab.~\ref{tab:scannet_semseg} shows the fine-tuning for semantic segmentation, in comparison with baseline pre-training approaches.
All pre-training methods significantly improve performance over training the semantic segmentation model from scratch. In particular, \OURS{} provides substantially better representation quality leading to a much stronger improvement over supervised ImageNet pre-training (+11 mIoU), and notably improving over MAE-unsupIN$\rightarrow$SN with ImageNet and ScanNet (+3.4 mIoU) and the 3D-based pre-training of Pri3D (+6.5 mIoU). We note that the multi-view 3D pre-training of Pri3D does not effectively embed informative 3D priors to ViT backbones, rather suffering from performance degradation from a ResNet-50 backbone.
In contrast, our \OURS{} pre-training can notably improve performance with a ViT backbone, indicating the effectiveness of our learned 3D priors.

\vspace{-0.35cm}
\paragraph{2D Object Detection and Instance Segmentation} We show that \OURS{} provides effective general 3D priors for a variety of image-based tasks, by evaluating downstream object detection and instance segmentation in Tab.~\ref{tab:scannet_det} and Tab.~\ref{tab:scannet_insseg}, respectively. Across all tasks, various pre-training approaches yield substantial improvement over training from scratch. Our masked 3D prior learning transfers effectively learned representations for object detection and instance segmentation, notably improving over the best-performing MAE-unsupIN$\rightarrow$SN (+4.3 AP@0.5 and +3.8 AP@0.5, respectively).

\begin{figure}
	\includegraphics[width=1.0\linewidth]{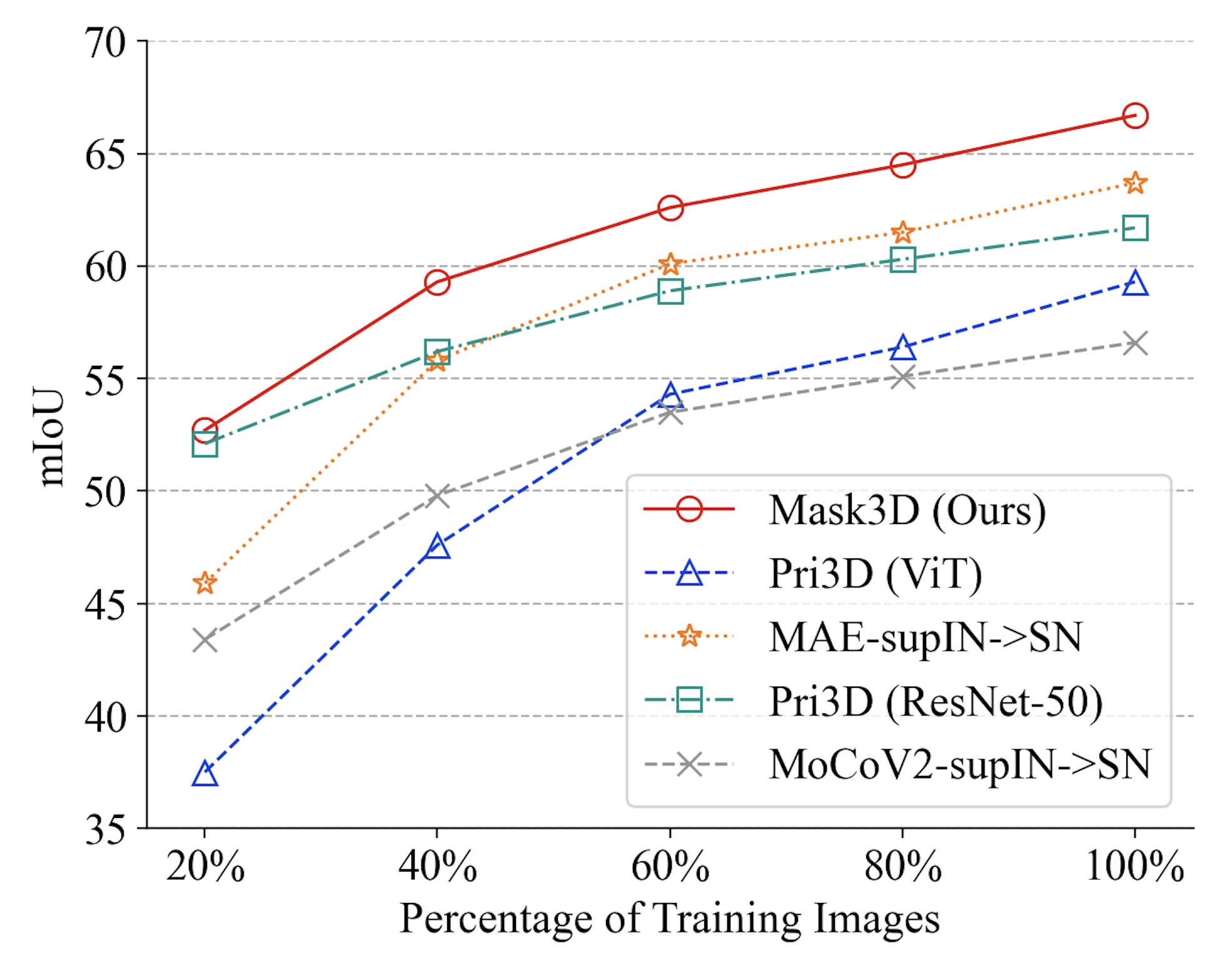}
 \vspace{-0.75cm}
	\caption{\textbf{Data-Efficient Results.} Compared to previous methods, \OURS{} demonstrates consistent improvements on ScanNet 2D semantic segmentation across a range of limited data scenarios. 
	\OURS{} is particularly effective for ViT pre-training, improving +$15.2\%$ mIoU over state-of-the-art Pri3D~\cite{hou2021pri3d} on a ViT backbone at 20\% of the training data.}
	\label{fig:data_efficient}
	\vspace{-0.2cm}
\end{figure}

\vspace{-0.35cm}
\paragraph{Data-Efficient Scenarios.}
We additionally show that our single-view RGB-D pre-training to embed 3D priors in limited data scenarios for downstream ScanNet semantic segmentation in Fig.~\ref{fig:data_efficient}.
\OURS{} shows consistent improvements across a range of limited data; even with only 20\% of the training data, we recover 80\% performance with 100\% training data available and improving +$15.2$ mIoU over Pri3D pre-training on a ViT backbone.

\subsection{NYUv2 Downstream Tasks}
\label{sec:nyuv2}
We demonstrate the generalizability of our 3D-imbued learned feature representations across datasets, using \OURS{} pre-trained on ScanNet and fine-tuned on NYUv2~\cite{silberman2012indoor} following the same fine-tuning setup as before. NYUv2 contains Microsoft Kinect RGB-D video sequences of indoor scenes, comprising 1449 densely labeled RGB images. We use the official 795/654 train/val split. Tables \ref{tab:nyu_semseg}, \ref{tab:nyu_det}, and \ref{tab:nyu_insseg} evaluate the downstream tasks of  2D semantic segmentation, object detection, and instance segmentation, respectively.
Across all three tasks on NYUv2 data, our \OURS{} pre-training achieves notably improved performance than training from scratch as well as the various baseline pre-training methods. In particular, we achieve an improvement of +6.9 mIoU, +14.1 AP@0.5, and +11.9 AP@ 0.5 over the common supervised ImageNet pre-training on semantic segmentation, object detection, and instance segmentation.

\begin{table*}
\small
  \centering
  \begin{tabular}{|l|c|c|c|}
  \hline
Pre-training Method~~~~~~~~ &~~~~~Backbone~~~~~&~~~~~Pre-training Data~~~~~& ~~ mIoU ~~ \\
  \hline
   \textcolor{gray}{Scratch}~&\textcolor{gray}{ResNet-50}&  \textcolor{gray}{None}    &\textcolor{gray}{24.8}~~~                              \\
   ImageNet Pre-training (supIN)~&ResNet-50 & ImageNet   &50.0~~~  \\
   Supervised Pre-training~&ViT & ImageNet+ScanNet       &~~~~~55.5~\tiny{\textBF{\textcolor{gray}{($+$5.5)}}}~  \\ \hline
   MoCoV2-supIN$\rightarrow$SN~\cite{he2020momentum} ~&ResNet-50                  & ImageNet+ScanNet     &~~~~~47.6~\tiny{\textBF{\textcolor{gray}{($-$2.4)}}}~  \\ 
   Pri3D~\cite{hou2021pri3d}                         ~&ResNet-50                  & ImageNet+ScanNet     &~~~~~54.2~\tiny{\textBF{\textcolor{gray}{($+$4.2)}}}~    \\ 
   Pri3D~\cite{hou2021pri3d}                         ~&ViT                        & ImageNet+ScanNet     &~~~~~53.2~\tiny{\textBF{\textcolor{gray}{($+$3.2)}}}~    \\ 
   MAE-unsupIN$\rightarrow$SN~\cite{he2021masked}    ~&ViT                        & ImageNet+ScanNet     &~~~~~54.9~\tiny{\textBF{\textcolor{gray}{($+$4.9)}}}~    \\ \hline 
   \OURS{} (Ours)                                           ~&ViT                    & ImageNet+ScanNet &~~~~~\textBF{56.9}~\tiny{\textBF{\textcolor{darkgreen}{($+$6.9)}}}~ \\ \hline
  \end{tabular}
  \vspace{-0.15cm}
  \caption{\textbf{NYUv2 2D Semantic Segmentation.} \OURS{} significantly outperforms state-of-the-art pre-training approaches, demonstrating its effectiveness in transferring to different dataset characteristics.}
\label{tab:nyu_semseg}
  \vspace{-0.35cm}
\end{table*}

\begin{table}
\small
  \centering
  \begin{tabular}{|l|ccc|}
  \hline
   Pre-training Method &   AP@0.5 & AP@0.75  & AP~~~~~~~  \\ \hline
   \textcolor{gray}{Scratch}&\textcolor{gray}{21.3}~~~~~~~~&\textcolor{gray}{10.3}~~~~~~~~&\textcolor{gray}{9.0}~~~~~~~~\\
   ImageNet Pretrain (supIN)&29.9~~~~~~~~&17.3~~~~~~~~&16.8~~~~~~~~\\ \hline
   MoCoV2-supIN$\rightarrow$SN~\cite{he2020momentum}&30.1~\tiny{\textBF{\textcolor{gray}{(+0.20)}}}&18.1~\tiny{\textBF{\textcolor{gray}{(+0.80)}}}&17.3~\tiny{\textBF{\textcolor{gray}{(+0.50)}}}\\
   Pri3D~\cite{hou2021pri3d}            &33.0~\tiny{\textBF{\textcolor{gray}{(+2.10)}}}&19.8~\tiny{\textBF{\textcolor{gray}{(+2.60)}}}&18.9~\tiny{\textBF{\textcolor{gray}{(+2.10)}}}\\
   MAE-unsupIN$\rightarrow$SN~\cite{he2021masked}&40.3~\tiny{\textBF{\textcolor{gray}{(+10.4)}}}&24.5~\tiny{\textBF{\textcolor{gray}{(+7.20)}}}&23.2~\tiny{\textBF{\textcolor{gray}{(+6.40)}}}\\ \hline 
   \OURS{} (Ours)&\textBF{44.0}~\tiny{\textBF{\textcolor{darkgreen}{(+14.1)}}}&\textBF{28.3}~\tiny{\textBF{\textcolor{darkgreen}{(+6.40)}}}&\textBF{25.9}~\tiny{\textBF{\textcolor{darkgreen}{(+9.10)}}}\\ \hline
  \end{tabular}
  \vspace{-0.25cm}
  \caption{\textbf{NYUv2 2D Object Detection.} Fine-tuning with \OURS{} pre-trained models leads to improved object detection results across different metrics, showing an effective transfer across dataset characteristics.
  }
  \vspace{-0.75cm}
\label{tab:nyu_det}
\end{table}

\begin{table}
\small
  \centering
  \begin{tabular}{|l|c|c|}
  \hline
Pre-training Method &Backbone&  mIoU  \\
  \hline
   ImageNet Pre-training (supIN)&ResNet-50 &~~~54.1~~~  \\ 
   Pri3D~\cite{hou2021pri3d}&ResNet-50 &~~~~~~~55.1~\tiny{\textBF{\textcolor{gray}{(+1.00)}}}\\ 
   MAE-unsupIN$\rightarrow$SN~\cite{he2021masked}&ViT&~~~~~~~64.7~\tiny{\textBF{\textcolor{gray}{(+10.6)}}}\\ \hline 
   \OURS{} (Ours)&ViT&~~~~~~~\textBF{66.4}~\tiny{\textBF{\textcolor{darkgreen}{(+12.3)}}} \\ \hline
  \end{tabular}
  \vspace{-0.25cm}
  \caption{\textbf{Cityscapes 2D Semantic Segmentation.} \OURS{} significantly outperforms state-of-the-art Pri3D as well as a strong MAE-style pre-training.
  This demonstrates the effectiveness of the transferability of \OURS{}, even under a significant domain gap. 
  }
  \vspace{-0.55cm}
\label{tab:cityscapes_semseg}
\end{table}

\subsection{Out-of-domain Transfer}
\label{sec:cityscapes}
While \OURS{} concentrates on pre-training for improving indoor scene understanding, we further demonstrate the effectiveness of our \OURS{} pre-training for the out-of-domain transfer on outdoor data, such as Cityscapes~\cite{cordts2016cityscapes}. We use the official data split of 3000 images for training and 500 images for the test. To evaluate the transferability in such a large domain gap scenario, we fine-tune the pre-trained models for the 2D semantic segmentation task in Tab.~\ref{tab:cityscapes_semseg}. Our approach maintains performance improvement over baseline pre-training methods such as Pri3D (+11.3 mIoU) and MAE-unsupIN$\rightarrow$SN (+1.7 mIoU). This indicates an encouraging transferability of our learned representations and their applicability to a variety of scenarios. Please refer to the supplemental material for more out-of-domain transfer results on more generally distributed data, such as ADE20K~\cite{zhou2017scene}.

\subsection{Ablation Studies}
\label{sec:ablation}

\paragraph{Does the pre-training masking ratio matter?} We show how different masking ratios influence downstream task results in Tab.~\ref{tab:vit_ratios} on ScanNet semantic segmentation. We found a performance gain when masking more RGB values (keeping 20\%), which in combination with the heavy depth masking (keeping 20\%) leads to the best performance.

\begin{table}

\centering
\small
  \begin{tabular}{|c|c|c|}
  \hline
RGB Ratio &Depth Ratio&  mIoU  \\ \hline
   20.0\%               &~0.0\%          & 65.2    \\ 
   20.0\%               &20.0\%          & ~~~{\bf 66.7}~~~    \\ 
   20.0\%               &80.0\%          & ~~~65.5~~~    \\ 
   50.0\%               &20.0\%          & 65.9     \\ 
   50.0\%               &50.0\%          & 64.7 \\ 
   80.0\%               &20.0\%          & 64.8    \\ 
   80.0\%               &50.0\%          &  64.8   \\ 
   100.0\%               &0.0\%          &   64.6  \\ 
   100.0\%               &20.0\%          &  64.8   \\ 
   100.0\%               &100.0\%          & 64.5    \\ \hline 
  \end{tabular}
\vspace{-0.2cm}
  \caption{\textbf{Ablation Study of Masking Ratios.} on ScanNet 2D semantic segmentation. We mask out different ratios of RGB and depth patches, where the ratio indicates the percentage of kept patches. Refer to supplemental material for a full list.
  }
\vspace{-0.5cm}
\label{tab:vit_ratios}
\end{table}

\vspace{-0.25cm}
\paragraph{What about other ViT variants?} In our experiments, we use ViT-B as the meta-architecture. We show \OURS{} also works in other ViT variants, such as ViT-L (see Tab.~\ref{tab:vit_variants}), which exhibits a similar trend of improvements.

\vspace{-0.25cm}
\paragraph{Does the normalization in the reconstruction loss help?} We normalize the features when computing the reconstruction loss and observe an improvement of +0.8\% mIoU in the semantic segmentation task on ScanNet.

\begin{table}[h!]
\small
\centering
  \begin{tabular}{|c|c|c|}
  \hline
Method & Backbone & mIoU  \\ \hline
   Train from Scratch & ViT & 32.6     \\ 
   MAE    &ViT &37.1 \\
   Mask3D &ViT &42.2 \\ \hline
  \end{tabular}
  \vspace{-0.25cm}
  \caption{\textbf{Results on ScanNet Semantic Segmentation without ImageNet pre-training.} Similar trend is seen as ImageNet pre-training. The Gap gets larger compared to ImageNet pre-training.}
  \vspace{-0.55cm}
  \label{tab:noimagenet}
\end{table}

\vspace{-0.25cm}
\paragraph{Compared to a pure depth prediction baseline.} In Tab.~\ref{tab:vit_ratios}, we demonstrate a superior performance with a 20\% kept patches of RGB and depth, compared to a pure depth prediction method (66.7 vs. 64.6). Note in the table, pre-training with 100\% RGB ratio and 0\% depth ratio is equivalent to a pure depth prediction from a RGB image.

\vspace{-0.25cm}
\paragraph{Color + depth reconstruction?} We found that having joint losses on color and depth during pre-training does not benefit performance (see the following Tab.~\ref{tab:rgbdepth}). The RGB reconstruction loss potentially makes pre-training easier, as we already have additional depth priors as guidance.

\begin{table}[h!]
\small
\centering
  \begin{tabular}{|c|c|c|}
  \hline
   Method & Reconstruction & mIoU  \\ \hline
   Mask3D & RGB+Depth  &  65.6\\
   Mask3D & Depth & 66.7 \\ \hline
  \end{tabular}
\vspace{-0.25cm}
  \caption{\textbf{ScanNet Semantic Segmentation.} RGB as an additional signal does not bring a significant improvement.}
  \label{tab:rgbdepth}
\vspace{-0.55cm}
\end{table}

\vspace{-0.25cm}
\paragraph{No Stage-I pre-training.} We observe a performance drop without ImageNet pre-training model as initialization for our pre-training. Since ImageNet pre-training is readily available and ScanNet has a relatively small amount of indoor data, we make ImageNet pre-training initialization as default, similar to Pri3D. Meanwhile, we conduct experiments without ImageNet pre-training in Tab.~\ref{tab:noimagenet}, and observe similar trends as when using ImageNet pre-training.

\begin{table}[h!]
\small
\centering
  \begin{tabular}{|c|c|c|}
  \hline
Datasets & Mask3D - Semantics & Mask3D  \\ \hline
ScanNet & 65.9 & 66.7 \\  
NYUv2 & 55.5 & 56.9 \\
CityScapes & 63.0 & 66.4 \\ \hline
\end{tabular}
\vspace{-0.25cm}
\caption{Semantic segmentation results (mIoU). ``Mask3D - Semantics" denotes pre-training using RGB+Semantics.}
\vspace{-0.2cm}
\label{tab:semantics}
\end{table}

\vspace{-0.3cm}
\paragraph{RGB + semantic segmentation as pre-training.} Using RGB and semantic segmentation for pre-training rather than depth completion achieved competitive results on ScanNet semantic segmentation, although this requires the use of semantic labels for the pre-training dataset, and is likely to be less transferable across domains than using depth completion. As shown in the following Tab.~\ref{tab:semantics}, the gap becomes larger when transferring to both NYUv2 and Cityscapes.

\begin{table}
\small
  \centering
  \begin{tabular}{|l|c|l|} \hline
Pre-training Method &Backbone&mIoU  \\ \hline
   Pri3D~\cite{hou2021pri3d}                         &ResNet-50   &60.2~~~~~~~    \\ \hline
   Pri3D~\cite{hou2021pri3d}                         &ViT-B       &59.3~\tiny{\textBF{\textcolor{gray}{(-0.9)}}}    \\ 
   MAE-unsupIN$\rightarrow$SN~\cite{he2021masked}    &ViT-B       &63.3~\tiny{\textBF{\textcolor{gray}{(+3.1)}}} \\
   \OURS{} (Ours)~                                   &ViT-B       &66.7~\tiny{\textBF{\textcolor{gray}{(+6.5)}}} \\ \hline
   Pri3D~\cite{hou2021pri3d}                         &ViT-L       &64.3~\tiny{\textBF{\textcolor{gray}{(+4.1)}}} \\ 
   MAE-unsupIN$\rightarrow$SN~\cite{he2021masked}    &ViT-L       &68.2~\tiny{\textBF{\textcolor{gray}{(+8.0)}}} \\ 
   \OURS{} (Ours)                                    &ViT-L       &\textBF{70.7}~\tiny{\textBF{\textcolor{darkgreen}{(+10.5)}}} \\ \hline
  \end{tabular}
  \vspace{-0.12cm}
  \captionof{table}{\textbf{ViT Variants on ScanNet 2D Semantic Segmentation.} \OURS{} yields consistent improvements for both ViT-B and ViT-L backbone architectures.}
  \vspace{-0.55cm}
\label{tab:vit_variants}
\end{table}
\section{Conclusion}
In this paper, we present \OURS{}, a new self-supervised approach to embed 3D priors into learned 2D representations for image scene understanding. We leverage existing large-scale RGB-D data to learn 3D priors without requiring any camera pose or multi-view correspondence information, instead learning geometric and structural cues through a pre-text reconstruction task from masked color and depth. We show that \OURS{} is particularly effective in pre-training the modern, powerful ViT backbones, with notable improvements across a variety of image-based tasks and datasets. We believe this shows the strong potential in effectively learning 3D priors and provides new avenues for such 3D-grounded representation learning.

%%%%%%%%% REFERENCES
{\small
\bibliographystyle{ieee_fullname}
\bibliography{egbib}

\begin{thebibliography}{10}\itemsep=-1pt

\bibitem{armeni_cvpr16}
Iro Armeni, Ozan Sener, Amir~R. Zamir, Helen Jiang, Ioannis Brilakis, Martin
  Fischer, and Silvio Savarese.
\newblock {3D} semantic parsing of large-scale indoor spaces.
\newblock In {\em ICCV}, 2016.

\bibitem{bachmann2022multimae}
Roman Bachmann, David Mizrahi, Andrei Atanov, and Amir Zamir.
\newblock Multimae: Multi-modal multi-task masked autoencoders.
\newblock In {\em Computer Vision--ECCV 2022: 17th European Conference, Tel
  Aviv, Israel, October 23--27, 2022, Proceedings, Part XXXVII}, pages
  348--367. Springer, 2022.

\bibitem{caron2021emerging}
Mathilde Caron, Hugo Touvron, Ishan Misra, Herv\'e J\'egou, Julien Mairal,
  Piotr Bojanowski, and Armand Joulin.
\newblock Emerging properties in self-supervised vision transformers.
\newblock In {\em Proceedings of the International Conference on Computer
  Vision (ICCV)}, 2021.

\bibitem{chang2017matterport3d}
Angel Chang, Angela Dai, Thomas Funkhouser, Maciej Halber, Matthias Niessner,
  Manolis Savva, Shuran Song, Andy Zeng, and Yinda Zhang.
\newblock Matterport3d: Learning from rgb-d data in indoor environments.
\newblock {\em arXiv preprint arXiv:1709.06158}, 2017.

\bibitem{chen2021empirical}
Xinlei Chen, Saining Xie, and Kaiming He.
\newblock An empirical study of training self-supervised vision transformers.
\newblock In {\em Proceedings of the IEEE/CVF International Conference on
  Computer Vision}, pages 9640--9649, 2021.

\bibitem{chen20214dcontrast}
Yujin Chen, Matthias Nie{\ss}ner, and Angela Dai.
\newblock 4dcontrast: Contrastive learning with dynamic correspondences for 3d
  scene understanding.
\newblock {\em arXiv preprint arXiv:2112.02990}, 2021.

\bibitem{choy20194d}
Christopher Choy, JunYoung Gwak, and Silvio Savarese.
\newblock {4D} spatio-temporal convnets: Minkowski convolutional neural
  networks.
\newblock In {\em CVPR}, 2019.

\bibitem{cordts2016cityscapes}
Marius Cordts, Mohamed Omran, Sebastian Ramos, Timo Rehfeld, Markus Enzweiler,
  Rodrigo Benenson, Uwe Franke, Stefan Roth, and Bernt Schiele.
\newblock The cityscapes dataset for semantic urban scene understanding.
\newblock In {\em CVPR}, 2016.

\bibitem{dahnert2021panoptic}
Manuel Dahnert, Ji Hou, Matthias Nie{\ss}ner, and Angela Dai.
\newblock Panoptic 3d scene reconstruction from a single rgb image.
\newblock {\em Advances in Neural Information Processing Systems}, 34, 2021.

\bibitem{dai2017scannet}
Angela Dai, Angel~X Chang, Manolis Savva, Maciej Halber, Thomas Funkhouser, and
  Matthias Nie{\ss}ner.
\newblock Scannet: Richly-annotated {3D} reconstructions of indoor scenes.
\newblock In {\em CVPR}, 2017.

\bibitem{dai20183dmv}
Angela Dai and Matthias Nie{\ss}ner.
\newblock 3dmv: Joint 3d-multi-view prediction for 3d semantic scene
  segmentation.
\newblock In {\em Proceedings of the European Conference on Computer Vision
  (ECCV)}, pages 452--468, 2018.

\bibitem{deng2009imagenet}
Jia Deng, Wei Dong, Richard Socher, Li-Jia Li, Kai Li, and Li Fei-Fei.
\newblock Imagenet: A large-scale hierarchical image database.
\newblock In {\em CVPR}, 2009.

\bibitem{devlin2018bert}
Jacob Devlin, Ming-Wei Chang, Kenton Lee, and Kristina Toutanova.
\newblock {BERT}: Pre-training of deep bidirectional transformers for language
  understanding.
\newblock In {\em NAACL}, 2019.

\bibitem{dosovitskiy2020image}
Alexey Dosovitskiy, Lucas Beyer, Alexander Kolesnikov, Dirk Weissenborn,
  Xiaohua Zhai, Thomas Unterthiner, Mostafa Dehghani, Matthias Minderer, Georg
  Heigold, Sylvain Gelly, et~al.
\newblock An image is worth 16x16 words: Transformers for image recognition at
  scale.
\newblock {\em arXiv preprint arXiv:2010.11929}, 2020.

\bibitem{el2021bootstrap}
Mohamed El~Banani and Justin Johnson.
\newblock Bootstrap your own correspondences.
\newblock In {\em Proceedings of the IEEE/CVF International Conference on
  Computer Vision}, pages 6433--6442, 2021.

\bibitem{engelmann20203d}
Francis Engelmann, Martin Bokeloh, Alireza Fathi, Bastian Leibe, and Matthias
  Nie{\ss}ner.
\newblock {3D-MPA: Multi-Proposal Aggregation for 3D Semantic Instance
  Segmentation}.
\newblock In {\em CVPR}, 2020.

\bibitem{graham20183d}
Benjamin Graham, Martin Engelcke, and Laurens van~der Maaten.
\newblock {3D} semantic segmentation with submanifold sparse convolutional
  networks.
\newblock In {\em CVPR}, 2018.

\bibitem{han2020occuseg}
Lei Han, Tian Zheng, Lan Xu, and Lu Fang.
\newblock {OccuSeg}: Occupancy-aware {3D} instance segmentation.
\newblock In {\em CVPR}, 2020.

\bibitem{he2021masked}
Kaiming He, Xinlei Chen, Saining Xie, Yanghao Li, Piotr Doll{\'a}r, and Ross
  Girshick.
\newblock Masked autoencoders are scalable vision learners.
\newblock {\em arXiv preprint arXiv:2111.06377}, 2021.

\bibitem{he2020momentum}
Kaiming He, Haoqi Fan, Yuxin Wu, Saining Xie, and Ross Girshick.
\newblock Momentum contrast for unsupervised visual representation learning.
\newblock In {\em CVPR}, 2020.

\bibitem{he2016deep}
Kaiming He, Xiangyu Zhang, Shaoqing Ren, and Jian Sun.
\newblock Deep residual learning for image recognition.
\newblock In {\em CVPR}, 2016.

\bibitem{hou20193d}
Ji Hou, Angela Dai, and Matthias Nie{\ss}ner.
\newblock {3D-SIS: 3D Semantic Instance Segmentation of RGB-D Scans}.
\newblock In {\em CVPR}, 2019.

\bibitem{hou2020revealnet}
Ji Hou, Angela Dai, and Matthias Nie{\ss}ner.
\newblock {RevealNet: Seeing Behind Objects in RGB-D Scans}.
\newblock In {\em CVPR}, 2020.

\bibitem{hou2020exploring}
Ji Hou, Benjamin Graham, Matthias Nie{\ss}ner, and Saining Xie.
\newblock Exploring data-efficient 3d scene understanding with contrastive
  scene contexts.
\newblock In {\em CVPR}, 2021.

\bibitem{hou2021pri3d}
Ji Hou, Saining Xie, Benjamin Graham, Angela Dai, and Matthias Nie{\ss}ner.
\newblock Pri3d: Can 3d priors help 2d representation learning?
\newblock In {\em ICCV}, 2021.

\bibitem{jiang2020pointgroup}
Li Jiang, Hengshuang Zhao, Shaoshuai Shi, Shu Liu, Chi-Wing Fu, and Jiaya Jia.
\newblock {PointGroup: Dual-Set Point Grouping for 3D Instance Segmentation}.
\newblock In {\em CVPR}, 2020.

\bibitem{lahoud20193d}
Jean Lahoud, Bernard Ghanem, Marc Pollefeys, and Martin~R Oswald.
\newblock 3d instance segmentation via multi-task metric learning.
\newblock In {\em ICCV}, 2019.

\bibitem{liu2021swin}
Ze Liu, Yutong Lin, Yue Cao, Han Hu, Yixuan Wei, Zheng Zhang, Stephen Lin, and
  Baining Guo.
\newblock Swin transformer: Hierarchical vision transformer using shifted
  windows.
\newblock In {\em Proceedings of the IEEE/CVF International Conference on
  Computer Vision}, pages 10012--10022, 2021.

\bibitem{nie2020rfd}
Yinyu Nie, Ji Hou, Xiaoguang Han, and Matthias Nie{\ss}ner.
\newblock Rfd-net: Point scene understanding by semantic instance
  reconstruction.
\newblock In {\em CVPR}, 2021.

\bibitem{voteNet}
Charles~R. Qi, Or Litany, Kaiming He, and Leonidas~J. Guibas.
\newblock Deep hough voting for {3D} object detection in point clouds.
\newblock {\em ICCV}, 2019.

\bibitem{qi2017pointnet}
Charles~R Qi, Hao Su, Kaichun Mo, and Leonidas~J Guibas.
\newblock Pointnet: Deep learning on point sets for {3D} classification and
  segmentation.
\newblock {\em CVPR}, 2017.

\bibitem{qi2017pointnetplusplus}
Charles~R Qi, Li Yi, Hao Su, and Leonidas~J Guibas.
\newblock Pointnet++: Deep hierarchical feature learning on point sets in a
  metric space.
\newblock {\em NeurIPS}, 2017.

\bibitem{radford2021learning}
Alec Radford, Jong~Wook Kim, Chris Hallacy, Aditya Ramesh, Gabriel Goh,
  Sandhini Agarwal, Girish Sastry, Amanda Askell, Pamela Mishkin, Jack Clark,
  et~al.
\newblock Learning transferable visual models from natural language
  supervision.
\newblock In {\em International Conference on Machine Learning}, pages
  8748--8763. PMLR, 2021.

\bibitem{silberman2012indoor}
Nathan Silberman, Derek Hoiem, Pushmeet Kohli, and Rob Fergus.
\newblock Indoor segmentation and support inference from {RGB-D} images.
\newblock {\em ECCV}, 2012.

\bibitem{song2015sun}
Shuran Song, Samuel~P Lichtenberg, and Jianxiong Xiao.
\newblock {SUN RGB-D: A RGB-D Scene Understanding Benchmark Suite}.
\newblock In {\em CVPR}, 2015.

\bibitem{thomas2019kpconv}
Hugues Thomas, Charles~R Qi, Jean-Emmanuel Deschaud, Beatriz Marcotegui,
  Fran{\c{c}}ois Goulette, and Leonidas~J Guibas.
\newblock {KPConv}: Flexible and deformable convolution for point clouds.
\newblock In {\em CVPR}, 2019.

\bibitem{wei2021masked}
Chen Wei, Haoqi Fan, Saining Xie, Chao-Yuan Wu, Alan Yuille, and Christoph
  Feichtenhofer.
\newblock Masked feature prediction for self-supervised visual pre-training.
\newblock {\em arXiv preprint arXiv:2112.09133}, 2021.

\bibitem{wu2019detectron2}
Yuxin Wu, Alexander Kirillov, Francisco Massa, Wan-Yen Lo, and Ross Girshick.
\newblock Detectron2.
\newblock \url{https://github.com/facebookresearch/detectron2}, 2019.

\bibitem{xie2020pointcontrast}
Saining Xie, Jiatao Gu, Demi Guo, Charles~R Qi, Leonidas~J Guibas, and Or
  Litany.
\newblock Pointcontrast: Unsupervised pre-training for {3D} point cloud
  understanding.
\newblock {\em ECCV}, 2020.

\bibitem{yang2019learning}
Bo Yang, Jianan Wang, Ronald Clark, Qingyong Hu, Sen Wang, Andrew Markham, and
  Niki Trigoni.
\newblock Learning object bounding boxes for {3D} instance segmentation on
  point clouds.
\newblock In {\em NeurIPS}, 2019.

\bibitem{yi2018gspn}
Li Yi, Wang Zhao, He Wang, Minhyuk Sung, and Leonidas Guibas.
\newblock {GSPN}: Generative shape proposal network for {3D} instance
  segmentation in point cloud.
\newblock In {\em CVPR}, 2019.

\bibitem{zhang2022pcr}
Yu Zhang, Junle Yu, Xiaolin Huang, Wenhui Zhou, and Ji Hou.
\newblock Pcr-cg: Point cloud registration via deep explicit color and
  geometry.
\newblock In {\em Computer Vision--ECCV 2022: 17th European Conference, Tel
  Aviv, Israel, October 23--27, 2022, Proceedings, Part X}, pages 443--459.
  Springer, 2022.

\bibitem{zhang2021self}
Zaiwei Zhang, Rohit Girdhar, Armand Joulin, and Ishan Misra.
\newblock Self-supervised pretraining of 3d features on any point-cloud.
\newblock In {\em Proceedings of the IEEE/CVF International Conference on
  Computer Vision}, pages 10252--10263, 2021.

\bibitem{zhang2020h3dnet}
Zaiwei Zhang, Bo Sun, Haitao Yang, and Qixing Huang.
\newblock H3dnet: 3d object detection using hybrid geometric primitives.
\newblock In {\em European Conference on Computer Vision}, pages 311--329.
  Springer, 2020.

\bibitem{zhou2017scene}
Bolei Zhou, Hang Zhao, Xavier Puig, Sanja Fidler, Adela Barriuso, and Antonio
  Torralba.
\newblock Scene parsing through ade20k dataset.
\newblock In {\em Proceedings of the IEEE conference on computer vision and
  pattern recognition}, pages 633--641, 2017.

\end{thebibliography}
}

%for arxiv:
\clearpage
\noindent \textbf{\Large Appendix} \\ 
\begin{appendix}
\section{More Qualitative Results}

\begin{figure*}[t]
	\centering
	\includegraphics[width=1.0\linewidth]{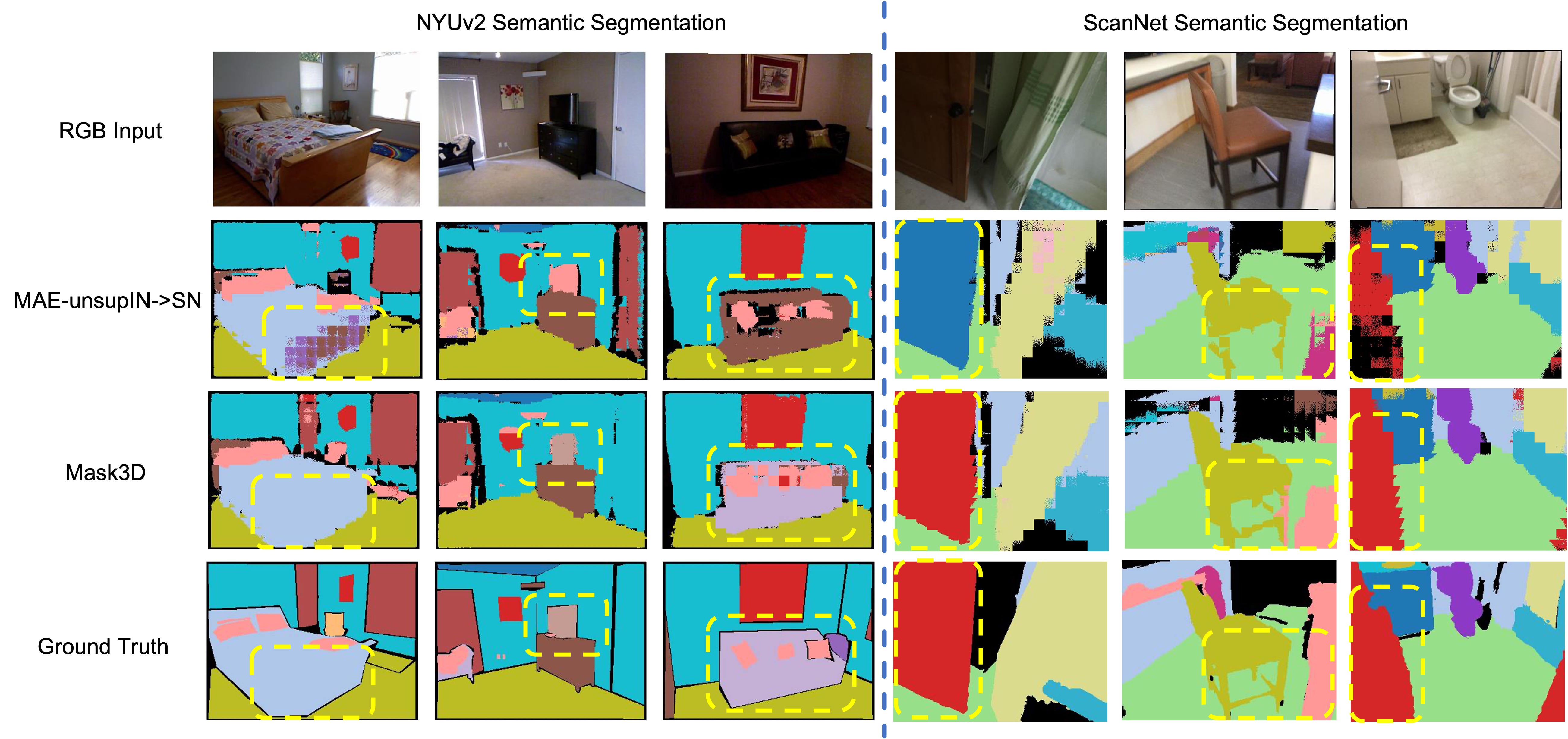}
	\caption{\textbf{Qualitative Results.} We show more visualizations on NYUv2 and ScanNet.
	}
	\label{fig:more_visuals}
\end{figure*}

In this section, we show more visualizations on NYUv2 and ScanNet semantic segmentation results across different methods in Figure~\ref{fig:more_visuals}.

\section{More Quantitative Results.}
In this section, we show more quantitative results, including the full list of ablation studies regarding different depth and RGB ratios. Next, we show the results using MAE unsupervised pre-trained model against supervised pre-trained checkpoint for Stage-I pre-training. Furthermore, we show another out-of-domain transfer learning experiment on ADE20K, a more generally distributed dataset.

\paragraph{Full List of RGB and Depth Ratios.} We show the expanded version of Table 8 below. We ablated different RGB and depth ratios, and found out that masking more RGB signal and bringing in sparse depth priors lead to higher mIoUs.

\begin{table}[h!]
\centering
  \begin{tabular}{|c|c|c|}
  \hline
RGB Ratio &Depth Ratio&  mIoU  \\ \hline
   20.0\%               &~0.0\%          & 65.2    \\ 
   20.0\%               &20.0\%          & {\bf 66.7} \\
   20.0\%               &50.0\%          & 66.4 \\
   20.0\%               &80.0\%          & 65.5    \\ 
   20.0\%               &100.0\%          &   65.3  \\ 
   50.0\%               &0.0\%          & 66.0    \\ 
   50.0\%               &20.0\%          & 65.9     \\ 
   50.0\%               &50.0\%          & 64.7 \\ 
   50.0\%               &80.0\%          & 65.4     \\ 
   50.0\%               &100.0\%          &  65.7   \\ 
   80.0\%               &0.0\%           &  64.4   \\ 
   80.0\%               &20.0\%          & 64.8    \\ 
   80.0\%               &50.0\%          &  64.8   \\ 
   80.0\%               &80.0\%          & 64.9    \\  
   80.0\%               &100.0\%          &  65.0   \\ 
   100.0\%               &0.0\%          &   64.6  \\ 
   100.0\%               &20.0\%          &  64.8   \\ 
   100.0\%               &50.0\%          &  64.5   \\ 
   100.0\%               &80.0\%          &  64.2   \\ 
   100.0\%               &100.0\%          & 64.5    \\ \hline 
  \end{tabular}
  \caption{\textbf{Full List of RGB and Depth Ratios} Results on ScanNet 2D semantic segmentation. We mask out different ratios of RGB and depth patches, where the ratio indicates the percentage of kept patches. }
\end{table}

\paragraph{Out-of-domain Transfer on ADE20K.} We observe a similar trend in the ADE20K dataset compared to the ScanNet, NYUv2 and Cityscapes (see the following Table~\ref{tab:ade20k}). We search for the best training recipes: learning rate 0.0001 with AdamW optimizer, training iterations 256k and batch size 16 on 8 GPUs.

\begin{table}[h!]
\small
\centering
  \begin{tabular}{|l|c|c|}
  \hline
   Pre-training Method & Backbone &  mIoU  \\ \hline
   MAE (MultiMAE reproduced)~\cite{bachmann2022multimae}  &  ViT & 46.2\\
   MAE (our reproduced)  &  ViT & 47.2\\
   MultiMAE~\cite{bachmann2022multimae} & ViT & 46.2 \\
   Mask3D & ViT & 47.7 \\ \hline
  \end{tabular}
  \caption{\textbf{Out-of-domain Transfer on ADK20k semantic segmentation}. Mask3D and MAE use the same training recipe for the downstream task, so it is a fair comparison. We can observe an improvement over MAE pre-trained checkpoint with masked depth priors for pre-training.}
  \label{tab:ade20k}
\end{table}

\paragraph{MAE-unsup-ViT vs. supIN-ViT.} We list the suggested baselines in the following Table~\ref{tab:vitbaselines}. We did not include supIN - ViT, since MAE-unsupIN - ViT shows a better performance than supIN - ViT (in the MAE paper), and MAE-unsupIN - ViT weights are readily available from official MAE code base whereas supIN ViT is not. Note that our method also uses MAE-unsupIN - ViT as initialization so it is a fair comparison, and this makes our method a pure unsupervised approach.

\begin{table}[h]
\small
\centering
  \begin{tabular}{|c|c|c|}
  \hline
Method & mIoU  \\ \hline
ViT Scratch & 32.6 \\ 
MAE-unsupIN - ViT & 63.7\\ \hline
Mask3D - ViT & 66.7 \\ \hline
  \end{tabular}
  \caption{\textbf{ViT baselines on ScanNet Semantic Segmentation.} A similar trend is observed in unsupervised setup.}
\label{tab:vitbaselines}
\end{table}

\paragraph{Limitations.}\label{sec:limitations} Our work aims to learn 3D geometric and spatial structures to benefit downstream scene understanding tasks. While we show that learning to reconstruct the dense depth can effectively embed learned geometric understanding, some geometric- and spatially-aware designs are not yet fully exploited, e.g., ViT-based multi-scale learning or exploring surface properties such as normals as a proxy loss.

\begin{figure}[h]
\includegraphics[width=1.0\linewidth]{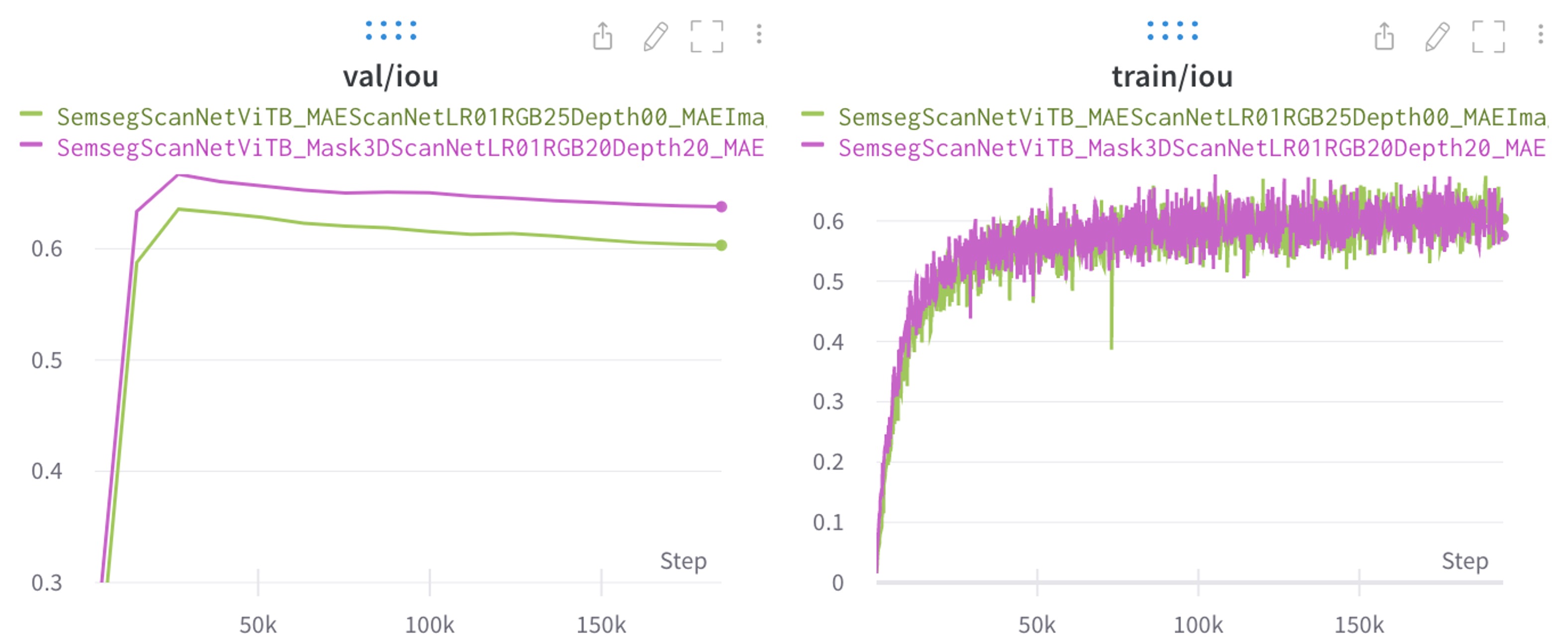}
\caption{\textbf{Training and Validation Curves.} A consistent gap is observed on ScanNet Semantic Segmentation between Mask3D and MAE-unsupIN$\rightarrow$SN.}
\label{fig:training_curves}
\end{figure}

\paragraph{Training and Validation Curves.} We demonstrate the training and validation curves of fine-tuning ScanNet Semantic Segmentation in Figure~\ref{fig:training_curves}. A consistent improvement can be observed from the curve.

\paragraph{Pre-training Orders.} We ablate the orders of pre-training in Table~\ref{tab:pretrain_order}.
 
\begin{table}[h]
\small
  \centering
  \begin{tabular}{|l|c|}
  \hline
Pre-training Orders& ~~ mIoU ~~ \\
  \hline
   MAE + Mask3D & 63.3\\
   MAE $\rightarrow{}$ Mask3D & 66.3 \\
   Mask3D $\rightarrow{}$ MAE & 63.1  \\ \hline
  \end{tabular}
  \caption{\textbf{ScanNet 2D Semantic Segmentation.} ``+'' indicates training together and ``$\rightarrow$'' indicates the pre-training order.}
\label{tab:pretrain_order}
\end{table}
\end{appendix}

\end{document}